\documentclass[conference]{IEEEtran}
\IEEEoverridecommandlockouts
\usepackage{cite}
\usepackage{amsmath,amssymb,amsfonts}
\usepackage{graphicx}
\usepackage{textcomp}
\usepackage{xcolor}

\usepackage{url}
\usepackage{hyperref}

\usepackage{booktabs}

\usepackage{multirow}
\usepackage{wrapfig}
\usepackage{cite}

\usepackage{setspace}

\newcommand{\ra}[1]{\renewcommand{\arraystretch}{#1}}

\usepackage{float}
\floatstyle{plaintop}
\restylefloat{table}

\usepackage{tablefootnote}

\usepackage{titlesec}
\titlespacing*{\section}{0pt}{0.1\baselineskip}{0.2\baselineskip}

\usepackage{subcaption}
\usepackage{lipsum}


\usepackage{algorithm,algpseudocode}% http://ctan.org/pkg/{algorithms,algorithmx}

\algnewcommand{\REQUIRE}[1]{%
  \State \textbf{Require:}
  \Statex \hspace*{\algorithmicindent}\parbox[t]{.8\linewidth}{\raggedright #1}
}

\algnewcommand{\Outputs}[1]{%
  \State \textbf{Output:}
  \Statex \hspace*{\algorithmicindent}\parbox[t]{.8\linewidth}{\raggedright #1}
}
\algnewcommand{\Initialize}[1]{%
  \State \textbf{Initialize:}
  \Statex \hspace*{\algorithmicindent}\parbox[t]{.8\linewidth}{\raggedright #1}
}
\algnewcommand{\algorithmicforeach}{\textbf{for each}}
\algdef{SE}[FOR]{ForEach}{EndForEach}[1]
  {\algorithmicforeach\ #1\ \algorithmicdo}% \ForEach{#1}
  {\algorithmicend\ \algorithmicforeach}% \EndForEach

\usepackage{fourier} 
\usepackage{array}
\usepackage{makecell}

\usepackage{scrextend}
\deffootnote[2em]{2em}{1em}{\textsuperscript{\thefootnotemark}\,}
\hypersetup{
    colorlinks=true,
    linkcolor=blue,
    filecolor=magenta,
    citecolor=violet,
    urlcolor=cyan,
    pdftitle={Overleaf Example},
    pdfpagemode=FullScreen,
    }
\usepackage{multicol}
\usepackage{lipsum}
\usepackage{mwe}

\usepackage{lscape}
\usepackage{graphicx}

\usepackage{authblk}

\def\BibTeX{{\rm B\kern-.05em{\sc i\kern-.025em b}\kern-.08em
    T\kern-.1667em\lower.7ex\hbox{E}\kern-.125emX}}
\begin{document}

\title{Decoupling Exploration and Exploitation for Unsupervised Pre-training with Successor Features
\thanks{This work was supported by the Australian Research Council through Discovery Early Career Researcher Award DE200100245 and Linkage Project LP210301046.}
}

%%\author{\IEEEauthorblockN{Anonymous Authors}}

\author[*1]{JaeYoon~Kim}
\author[$\dagger$1]{Junyu~Xuan}
\author[$\dagger$2]{Christy~Liang}
\author[$\dagger$1]{Farookh~Hussain\vspace{-1em}}
\affil[*]{Email:\; JaeYoon.Kim@student.uts.edu.au}
\affil[$\dagger$]{Email:\;\{Junyu.Xuan,\;Jie.Liang,\;Farookh.Hussain\}@uts.edu.au}
\affil[1]{Australian Artificial Intelligence Institute (AAII), University of Technology Sydney, Australia}
\affil[2]{Visualisation Institute, University of Technology Sydney, Australia}

\iffalse
\author{\IEEEauthorblockN{1\textsuperscript{st} Given Name Surname}
\IEEEauthorblockA{\textit{dept. name of organization (of Aff.)} \\
\textit{name of organization (of Aff.)}\\
City, Country \\
email address or ORCID}
\and
\IEEEauthorblockN{2\textsuperscript{nd} Given Name Surname}
\IEEEauthorblockA{\textit{dept. name of organization (of Aff.)} \\
\textit{name of organization (of Aff.)}\\
City, Country \\
email address or ORCID}
\and
\IEEEauthorblockN{3\textsuperscript{rd} Given Name Surname}
\IEEEauthorblockA{\textit{dept. name of organization (of Aff.)} \\
\textit{name of organization (of Aff.)}\\
City, Country \\
email address or ORCID}
\and
\IEEEauthorblockN{4\textsuperscript{th} Given Name Surname}
\IEEEauthorblockA{\textit{dept. name of organization (of Aff.)} \\
\textit{name of organization (of Aff.)}\\
City, Country \\
email address or ORCID}
\and
\IEEEauthorblockN{5\textsuperscript{th} Given Name Surname}
\IEEEauthorblockA{\textit{dept. name of organization (of Aff.)} \\
\textit{name of organization (of Aff.)}\\
City, Country \\
email address or ORCID}
\and
\IEEEauthorblockN{6\textsuperscript{th} Given Name Surname}
\IEEEauthorblockA{\textit{dept. name of organization (of Aff.)} \\
\textit{name of organization (of Aff.)}\\
City, Country \\
email address or ORCID}
}
\fi
\maketitle

\begin{abstract}
\iffalse
This document is a model and instructions for \LaTeX.
This and the IEEEtran.cls file define the components of your paper [title, text, heads, etc.]. *CRITICAL: Do Not Use Symbols, Special Characters, Footnotes, 
or Math in Paper Title or Abstract.
\fi
Unsupervised pre-training has been on the lookout for the virtue of a value function representation referred to as successor features (SFs), which decouples the dynamics of the environment from the rewards. It has a significant impact on the process of task-specific fine-tuning due to the decomposition. However, existing approaches struggle with local optima due to the unified intrinsic reward of exploration and exploitation without considering the linear regression problem and the discriminator supporting a small skill sapce. We propose a novel unsupervised pre-training model with SFs based on a non-monolithic exploration methodology. Our approach pursues the decomposition of exploitation and exploration of an agent built on SFs, which requires separate agents for the respective purpose. The idea will leverage not only the inherent characteristics of SFs such as a quick adaptation to new tasks but also the exploratory and task-agnostic capabilities. Our suggested model is termed Non-Monolithic unsupervised Pre-training with Successor features (NMPS), which improves the performance of the original monolithic exploration method of pre-training with SFs. NMPS outperforms Active Pre-training with Successor Features (APS) in a comparative experiment.

\end{abstract}


\section{Introduction} \label{introduction}
%%This document is a model and instructions for \LaTeX.
%%Please observe the conference page limits. 
Unsupervised pre-training leverages previously acquired behaviour without requiring an extrinsic reward, enabling an agent to rapidly adapt to new downstream tasks during fine-tuning. It has contributed significantly to Computer Vision (CV) \cite{116} and Natural Language Processing (NLP) \cite{112}, overcoming the weak point of supervised reinforcement learning which is limited to a single task.

As skill-based unsupervised reinforcement learning (RL) has been researched, one of current mainstream unsupervised pre-training RLs involves an agent maximizing the mutual information between an encoded state and a policy skill to learn an explicit skill vector. This approach takes advantage of algorithms such as VIC \cite{56}, DIAYN \cite{55}, and VALOR \cite{57}. However, during fine-tuning, these algorithms may suffer from issues related to inferior generalization and sluggish task inference due to the inefficient interpolation relying on latent behaviour spaces learnt during unsupervised pre-training.

To address these issues, research on the value function using Successor Features (SFs) has been conducted to decouple the dynamics of environment from the reward \cite{73, 102}. Using SFs for value function approximation makes it possible to relearn only the applicable component in case of changes to dynamics or reward. VISR \cite{100} takes advantage of SFs focusing on 'task inference' for fine-tuning after it is pre-trained. Although it has strengths against VIC and DIAYN in terms of certain issues, it also has the weakness of inefficient exploration.

\begin{figure}
 \vspace*{-5mm} 
 \centering

 \includegraphics[scale=0.2]{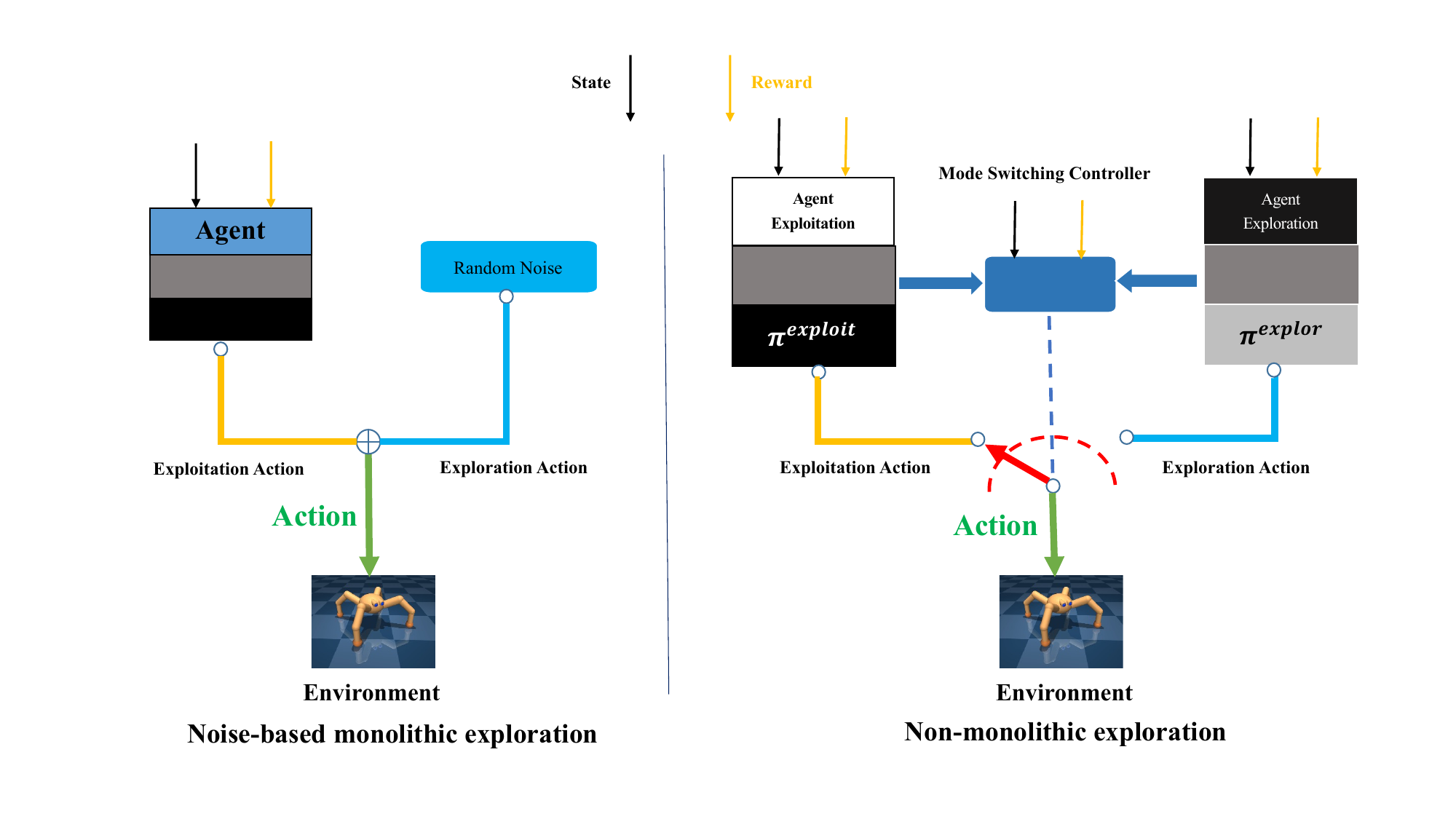}
 \vspace*{-5mm}
 \caption{An example of noise-based monolithic exploration (left) and non-monolithic exploration (right). In the noise-based monolithic exploration, noise and the agent's action play roles in exploration and exploitation, respectively. The final action, 'Action', taken by the agent in the environment is the result of adding the action with a noise. In the non-monolithic exploration, the exploitation agent  and the exploration agent work for their own purposes with the help of a mode-switching controller. The mode-switching controller considers the state of one or both agents.}
 \label{fig:nonmono1}
\end{figure}

APS \cite{81}, one of the competence-based algorithms mentioned in \cite{84},  utilizes combined intrinsic reward with rewards of VISR and Active Pre-Training (APT) \cite{107}, allowing `task inference' and `exploration', respectively. Even though the mutual information exploration method of APS is considered a recent leading method, its performance is not sufficient since its combined intrinsic reward interferes with the original intention of the respective intrinsic reward from VISR or APT. In addtion, the competence-based approaches such as APS have a chronic issue supporting small skill spaces due to the requirement of a precise discriminator \cite{84}. The linear reward-regression problem mentioned in \cite{100} gets worse since these interference can ultimately cause damage to the role of classifier or discriminator for fine-tuning the pre-trained RL agent.

In our research, we propose a novel unsupervised pre-training model for the agent based on SFs ensuring the decomposition of `task inference' and `exploration' to improve performance. In the methodology, the agent that conducts `task inference' should primarily focus on the `exploitation' without `exploration'. To avoid a single agent that conducts both `exploitation' and `exploration', there should be discrete agents responsible for each task.

Since a RL agent built on noise-based monolithic exploration typically is not suitable for our research,  we focus on separating `exploitation' and `exploration' using a non-monolithic exploration methodology, as shown in Fig.\ref{fig:nonmono1}. There should be several answers to build a decoupled exploration methodology that allows an agent using SFs to switch between two modes, exploitation and exploration, during unsupervised pre-training. The questions that arise are: How can such a methodology be conducted for unsupervised pre-training of SFs to switch an exploration mode? How can separate agents be trained during the pre-training of the decoupled exploration methodology? Is it possible for an agent using SFs to overcome a chronic issue supporting small skill spaces resulting from the requirement of a precise discriminator?  The contributions of our research are three-folded:

\begin{itemize} 
\item {The development of a decoupled exploration methodology based on non-monolithic exploration methodology to improve the performance of a monolithic agent using successor features during unsupervised pre-training.}
\item {An optimized training method of separate agents  during the pre-training phase of decoupled exploration methodology for the agent using successor features.}
\item  {The ability taking advantage of a competence-based algorithm as a separate exploration agent for greater flexibility and generalization of discriminator of exploitation agent.}
\end{itemize}

\iffalse
This research is explained in the following sections: Section \ref{Preliminary knowledge} observes SFs and non-monolithic exploration; Section \ref{Related work} reviews related research; Section \ref{Our methodology} explains our proposed decoupled exploration methodology; Section \ref{Experiments} describes the experiments conducted to measure the performance of our decoupled exploration methodology during the fine-tuning phase after unsupervised pre-training phase; in Section \ref{Discussion}, we discuss various issues identified in the experiments; and finally, we conclude this research and suggest future work in Section \ref{Conclusion}.
\fi

\begin{table}[!t]
   \caption{Key Notations.}
   \begin{minipage}{\columnwidth}
     \begin{center}
       %%\begin{tabular}{lp{0.65\columnwidth}}
       \begin{tabular}{lp{0.55\columnwidth}}
         \toprule
         Symbol & Meaning \\
         \hline
         \(t\) & action step\\
         \(s\) & current state\\    
         \(s'\) & next state\\          
         \(a\) & The final action to an environment\\        
         \(a^{exploit}_{t}\) & The action of $\pi^{exploit}$\\
         \(a^{explor}_{t}\) & The action of $\pi^{explor}$\\
         \(\textbf{z}\) & Encoded observation (or state)\\ 
         \(\textbf{w}\) & Task vector\\ 
         \(r^{exploitation}\;or\;r^{exploit}\) & Intrinsic reward of APS exploitation\\
         \(r^{exploration}\;or\;r^{explor}\) & Intrinsic reward of APS exploration\\    
         \(Exploit\) & The agent of exploitation\\   
         \(Explor\) & The agent of exploration\\
         \(\pi^{exploit}\) & The policy of Exploit\\   
         \(\pi^{explor}\) & The policy of Explor\\
         \(\phi^{exploit}\) & The feature of Exploit\\   
         \(\phi^{explor}\) & The feature of Explor\\
         \(Q^{\text{exploit}}\) & The state-action value function of Exploit\\
         \(Action\) & The action, chosen by Homeo, to an environment\\
         \(D_{promise}(t-k,t)\) & The value promise discrepancy\\
         \(D_{promise}^{explot}\) & The value promise discrepancy of $\pi^{exploit}$ \\         
         \(f_{homeo}\) & The function of homeostasis of the policy of Exploit, $\pi^{exploit}$\\
         \(\rho\) & The preset value of target rate, i.e. the average number of switches of the reference model, \textcolor{black}{which is selected among (0.1; 0.01; 0.001; 0.0001) for pre-training}\\
         \(D_{R}^{\#}\) & The data of a common replay buffer and its encoded state through the state encoder of $\#$ which is Exploit or Explor\\
         %%\(Homeo\) & The function of Homeostasis \\
         \bottomrule
       \end{tabular}
     \end{center}
   \end{minipage}
   \label{notation}
\end{table}

\section{PRELIMINARY KNOWLEDGE} \label{Preliminary knowledge}
\subsection{Successor features}
%\cite{104} claims a value function can be the product of a %reward array and a successor matrix in a tabular domain as %follows. 

%\begin{normalsize }
%    \begin{equation}
%        V(s) = R(s) M(s, s')
%    \end{equation}
%\end{normalsize }
%where R(s) is a reward array, M(s, s') is a successor matrix %which is a successor representation and V(s) is a value function.

\cite{117} further extends the successor representation of \cite{104} to a continuous space as
%%%\begin{multline}
\begin{flalign}
\label{eqn:linear_regresssion}
    \begin{aligned}
    r (s, a_t, s') &= \phi (s, a, s')^T \textbf{w} \\ 
    Q^{\pi}(s, a) &= E^{\pi} \biggl[ \sum^{\infty}_{i=t}\gamma^{i-t}\phi_{i+1}\;|\;S_{t} = s,\;A_{t} = a \biggl]^T \textbf{w} \\ 
    &= \psi^{\pi} (s, a)^T \textbf{w}
    %\shoveleft{\qquad \quad \: \; \, \equiv \psi^{\pi} (s, a)^T %\textbf{w} \hfill}
    \end{aligned}
\end{flalign}    
%%%\end{multline}
where \textbf{w} is a task vector, $r (s, a_t, s')$ is a reward function, s is a state and $\phi_t = \phi (S_{t}, A_{t}, S^{'}_{t})$ and $\psi^{\pi} (s, a)$ are a successor feature of policy $\pi$. Generalized Policy Improvement (GPI) based on SFs has also been claimed to have a transfer reinforcement learning framework \cite{99, 101}.

\begin{figure*}
  \vspace*{-10mm} 
  \centering
  \includegraphics[scale=0.35]{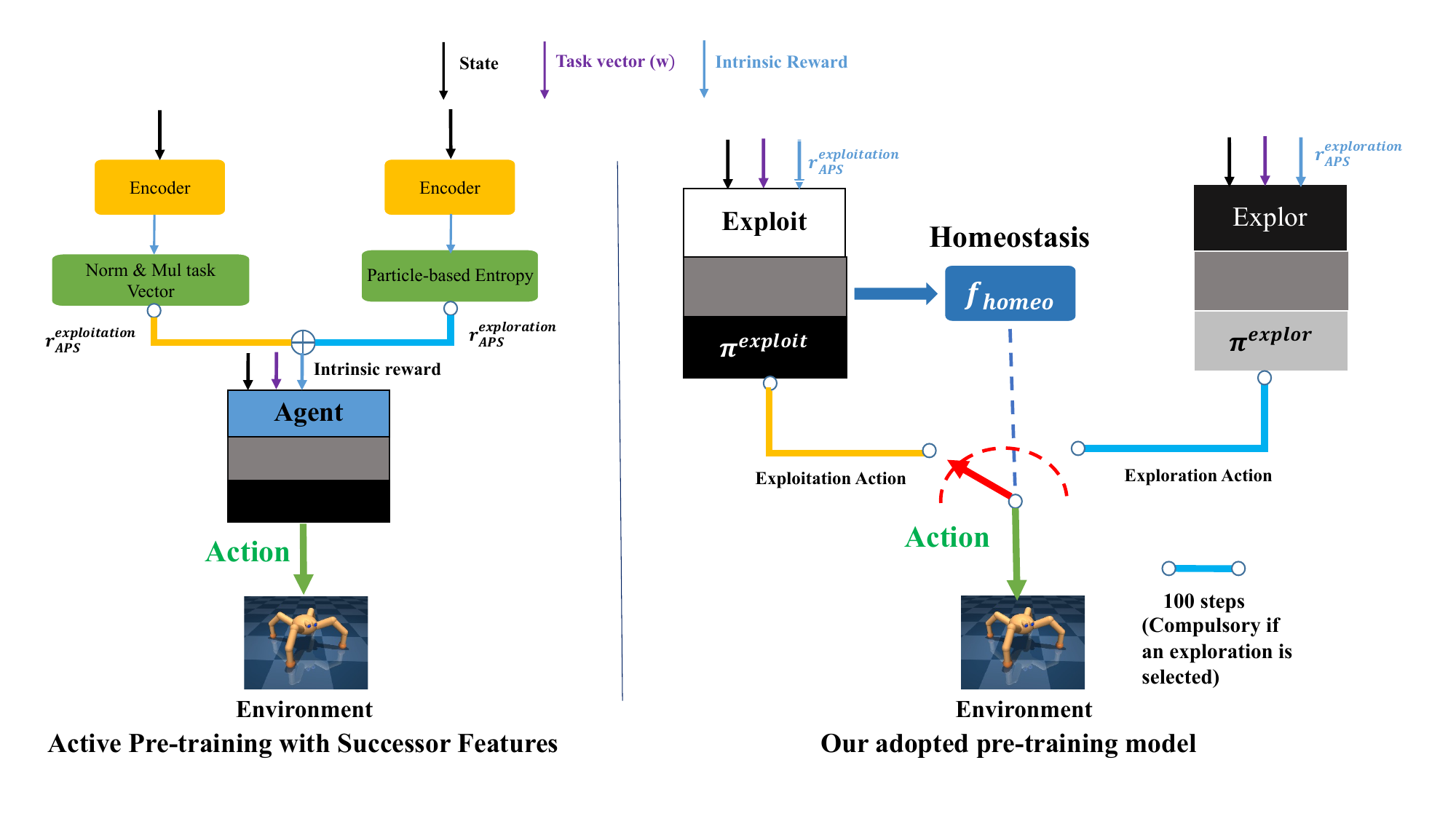}
  %%\hspace*{-0.4cm}
  \vspace*{-5mm}
  \caption{The architecture of our proposed pre-training model, NMPS, (right) compared with Active Pre-training with Successor Features  (left)  }
  \label{fig:both_architecture}
\end{figure*}

\subsection{Non-monolithic exploration}
As described in \cite{127, 128}, \cite{63} claims the importance of `when' to explore, which is a non-monolithic exploration approach, as opposed to `how' to explore which is typically associated with monolithic exploration. As its switching method, it adopts `Homeostasis' \cite{64}, which is abbreviated as 'Homeo', using the difference of value function between k steps which is called the `value promise discrepancy' as
\vspace*{-3mm}
\begin{equation}
\label{eq:valuepromise}
D_{promise}(t-k,t) := \; \big | V(s_{t - k}) - \sum_{i=0}^{k-1}\gamma^{i}R_{t-i}  -  \gamma^{k}V(s_{t}) \big |
\end{equation}
where $V(s)$ denotes the agent's value estimate at state s, $R$ denotes the reward and $\gamma$ denotes a discount factor.

\section{Related work} \label{Related work}

%%\subsection{The contradiction of intrinsic reward of APS}

%%%\cite{124} \cite{119} \cite{120} \cite{121} \cite{122}

\iffalse
Our research combines following three fields: The first is SFs. The second is the unsupervised behaviour learning algorithm using intrinsic behaviour learning to maximize entropy \cite{120, 121} or mutual information \cite{ 100, 123}. The last is the non-monolithic exploration methodology. This section reviews SFs and the unsupervised pre-training algorithm to maximize entropy or mutual information.
\fi

The research in \cite{94} makes use of SFs to transfer previous knowledge to similar environments in simulated and real experiments. As the importance of SFs is magnified, \cite{118} researches visual semantic planning to predict a sequence of actions for a desired goal state by using a deep successor representation. \cite{96} shows that although SFs in transfer learning promote learning speed, there is a restriction in transferring knowledge between tasks when an SF representation relies on the task's optimal policy. UVFAs, which is $V(s,g,\theta)$ using parameters $\theta$, claims the importance of value function approximation of adjacent states as well as adjacent goals \cite{86}. USFA \cite{73} combines UVFAs, SFs and GPI to support multi-task deep reinforcement learning.

In order to overcome 'linear regression problem', VISR takes advantage of SFs and the behavioral mutual information (BMI) \cite{ 55, 106}, which constructs a reward function as \cite{100} to support fast task inference as follows:
\vspace*{-2mm}
\begin{normalsize}
    \begin{equation}
        \log q (w | s) = \phi (s)^T \textbf{w}
    \end{equation}
\end{normalsize} where q is a BMI discriminator.

Since APS makes use of the strong points of both VISR  and APT, the intractable conditional entropy $-H(s|z)$ can be lower bounded. For the variational lower bound of APS, the intrinsic reward for the pre-training is the sum of $\text{log} q (w | s)$ of VISR and $r_{t}^{APT}(s)$ of APT.
%\begin{flalign}
%\begin{equation}
\begin{align}
\begin{split}
    %\begin{aligned}
    %\begin{equation}
    \label{aps_equation}
    %\begin{aligned }
    r_{APS}(s, a, s' ) &= r_{APS}^{exploitation}(s, a, s') + r_{APS}^{exploration}(s, a, s')\\
    &= \phi (s)^T w + \log  \Biggl(1 + \frac{1}{k} \sum_{h^{(j)} \in N_{k}(h)} \big\| h - h^{(j)} \big\|_{n_h}^{n_h} \Biggr)\\
    &= \log q (w | s) + r_{t}^{APT}(s)
    %%\\ where \hfill h = \phi (s')
    %%\\ & where \;  h = \phi (s') and N_{k}() denotes the k nearest neighbors               
    %\end{aligned}
    %\end{equation}
    %\end{aligned}
\end{split}
\end{align}    
%\end{flalign}
%\end{equation}
where $h = \phi (s')$ and $N_{k}(\cdot)$ denotes the k nearest neighbors.

\begin{table*}[t]
%%\vspace*{-20mm} 
\centering
\ra{1.0}
\begin{tabular}{@{}rrrrcrrrcrrr@{}}\toprule
%%%& \multicolumn{2}{c}{$The\;reward\;of\;Agent$} & \phantom{abc}& %%%\multicolumn{2}{c}{$Replay\;Buffer$} &
%%%\phantom{abc} & \multicolumn{2}{c}{$The\;train\;of\;feature$}\\

& \multicolumn{2}{c}{Reward} & \phantom{a}& \multicolumn{2}{c}{$D_{R}^{\#}$} & \phantom{a} & \multicolumn{2}{c}{T. f or s} & \phantom{a} & \phantom{abc} \\

\cmidrule{2-3} \cmidrule{5-6} \cmidrule{8-9}
The variants of NMPS & Exploit & Explor  && Exploit & Explor && Et & Er && Action\\ \midrule
%%%%$dir=1$\\

$NMPS\_X\_sep^{ex}$ & \multirow{12}{*}{$r^{exploit}$} & $r^{explor}$ && $D_{R}^{exploit}$ & $D_{R}^{explor}$  && \checkmark & \checkmark && Homeo \\
$NMPS\_X\_sep^{e*}$ &  & $r^{explor}$  && $D_{R}^{exploit}$ & $D_{R}^{explor}$  && \checkmark & x && Homeo \\
%%%$NMPS\_X\_sep^{*x}$ &  & $r^{explor}$  && $D_{R}^{exploit}$ & $D_{R}^{explor}$ && x & \checkmark && Homeo \\

$NMPS\_X\_exploit^{ex}$ &  & $r^{explor}$ && $D_{R}^{exploit}$ & common && \checkmark & \checkmark && Homeo\\
$NMPS\_X\_exploit^{e*}$ &  & $r^{explor}$ && $D_{R}^{exploit}$ & common && \checkmark & x && Homeo\\
%%%$NMPS\_X\_exploit^{*x}$ &  & $r^{explor}$ && $D_{R}^{exploit}$ & common && x & \checkmark && Homeo\\

$NMPS\_X\_explor^{ex}$ &  & $r^{explor}$ && common & $D_{R}^{explor}$ && \checkmark & \checkmark && Homeo\\
$NMPS\_X\_explor^{e*}$ &  & $r^{explor}$ && common & $D_{R}^{explor}$ && \checkmark & x && Homeo\\
%%%$NMPS\_X\_explor^{*x}$ &  & $r^{explor}$ && common & $D_{R}^{explor}$ && x & \checkmark && Homeo\\

$NMPS\_D\_sep^{ex}$ &  & DIAYN && $D_{R}^{exploit}$ & $D_{R}^{DIAYN}$ && \checkmark & \checkmark && Homeo\\
$NMPS\_D\_sep^{e*}$ &  & DIAYN && $D_{R}^{exploit}$ & $D_{R}^{DIAYN}$ && \checkmark & x && Homeo\\
%%$NMPS\_D\_sep^{*x}$ &  & DIAYN && $D_{R}^{exploit}$ & $D_{R}^{DIAYN}$ && x & \checkmark && Homeo\\

$NMPS\_D\_sep^{ex\_D}$ &  & DIAYN && $D_{R}^{exploit}$ & $D_{R}^{DIAYN}$ &&  \checkmark & \checkmark && DIAYN\\

$NMPS\_D\_sep^{e*\_D}$ &  & DIAYN && $D_{R}^{exploit}$ & $D_{R}^{DIAYN}$ && \checkmark & x && DIAYN\\

$NMPS\_D\_sep^{ex\_D\_A10}$ &  & DIAYN && $D_{R}^{exploit}$ & $D_{R}^{DIAYN}$ && \checkmark & \checkmark && DIAYN\\
$NMPS\_D\_sep^{e*\_D\_A10}$ &  & DIAYN && $D_{R}^{exploit}$ & $D_{R}^{DIAYN}$ && \checkmark & x && DIAYN\\
%%%$NMPS\_D\_explor^{e*}$ &  & DIAYN && common & $D_{R}^{DIAYN}$  &&  \checkmark & x && Homeo\\
%%%$NMPS\_D\_explor^{*x}$ &  & DIAYN && common & $D_{R}^{DIAYN}$  && x & \checkmark && Homeo\\

%%%$NMPS\_D\_sep^{w}$ &  & DIAYN && $D_{R}^{exploit}$ & $D_{R}^{DIAYN}$ && \checkmark & \checkmark && DIAYN\\
\bottomrule

\end{tabular}
\caption{\textcolor{black}{Comparing all variants of NMPS based on 4 factors that are `Reward', `$D_{R}^{\#}$', `The train of feature or skill ('T. f or s' marked in the table)' and `Action' on the model notation of $NMPS\_\bigl\{X\;or\;D\bigr\}\_\bigl\{sep,\;exploit\;or\;explor\bigr\}\_\bigl\{e,\;x\;or\;*\bigr\}$.}}
\label{model_notation}
\end{table*}

%%%%%%%%%%%%%%%%%%%%%%%%%%%%%%%%%%%%%%%%%%%%%%%%%%%%%%%%%%%%%%%%%%%%%%%%
\section{Our methodology} \label{Our methodology}

An agent using SFs computes $\psi^{\pi} (s, a)$  during unsupervised pre-training. Then,
during supervised fine-tuning, the agent can find a $\textbf{w} \in R^{d}$  by solving a regression problem through $r (s, a_t, s')$ of Eq. (\ref{eqn:linear_regresssion}). Afterward, it  computes based on $Q^{\pi}(s, a)$ of Eq. (\ref{eqn:linear_regresssion}). However, APS cannot exactly solve a regression problem during supervised fine-tuning with  $\psi^{\pi} (s, a)$ computed based on $r_{APS}(s, a, s' )$ of Eq. (\ref{aps_equation}) during unsupervised pre-training.

The discrete agents using SFs, which are Exploit and Explor, are required to divide the united intrinsic reward of APS, which is Eq. (\ref{aps_equation}), to overcome an implicit issue that violates the linear reward-regression problem of Eq. (\ref{eqn:linear_regresssion}) during task-specific fine-tuning. After pre-training, Exploit is used for fine-tuning and  Explor is discarded. 

Since our model utilizes discrete agents, we explore various model variants to enhance its performance across different domains. Table \ref{model_notation} shows all evaluated models, which are composed of 4 independent factors, to find the optimal model in each domain. There are 4 groups: $NMPS\_X\_sep$, $NMPS\_X\_exploit$, $NMPS\_X\_explor$, $NMPS\_D\_sep$ in NMPS model. Among them, the standard type is $NMPS\_X\_sep^{ex}$. 
\begin{itemize} 
\item In the `Exploit' of `Reward' column, the intrinsic rewards of exploitation agent which is abbreviated as 'Exploit' are $r^{explitation}_{APS}$ of APS. In the `Explor' of `Reward' column, the intrinsic rewards of exploration agent which is abbreviated as 'Explor' are $r^{exploration}_{APS}$ of APS or the reward of DIAYN with X and D of model notation, respectively. DIAYN in `Reward' also means that Explor is DIAYN.

\item `common' in `$D_{R}^{\#}$' means that the active $D_{R}^{\#}$ is used for both agents in common. Furthermore, `sep(erate)', `exploit' and `explor' of model notation means the use of $D_{R}^{\#}$ of both agents, only $D_{R}^{exploit}$ of exploitation agent and only $D_{R}^{explor}$ of exploration, respectively. 

\item `\checkmark' and `x' in `T. f or s' means that the feature or skill of Explor is trained or not. `e', `x' or `*' of model notation means that the feature of Exploit or the feature or skill of Explor is trained or not.

\item `Homeo' in `Action' means that the final action is determined in Exploit or Explor depending on the decision of Homeo. `DIAYN' in `Action' with `D' of notation means that the only exploration agent of $NMPS\_D\_sep^{ex}$, DIAYN, decides Action.  $NMPS\_D\_sep^{\#\#\_D}$ denotes that the default dimensions of feature of Exploit and skill of Explor are 10 and 16, respectively. 'A10' in $NMPS\_D\_sep^{\#\#\_D\_A10}$ denotes that all dimensions of feature of Exploit and skill of Explor are 10.
\end{itemize}
%\vspace*{-7mm}
The details of four independent factors are explained in the following section.

\subsection{The model with a non-monolithic exploration}
We adopt the non-monolithic model used in \cite{63} as a reference method inspiring our model, as shown in  Fig.\ref{fig:both_architecture} for the decoupled exploration methodology that our research seeks. In our adopted pre-training model, the decision of Action is decided by Homeo  using the value promise discrepancy of Exploit, based on $Q^{\text{exploit}}$, as an input:
\vspace*{-2mm}
\begin{equation}
\begin{split}
%%\begin{multline}
%%\[
%%Action = \begin{array}{lr}
D_{\text{promise}}^{\text{exploit}} & = D_{\text{promise}}^{\text{exploit}}(t-k,t)\\
Action & = \begin{cases}
    a_t^{exploit} \sim \pi^{exploit}, & \text{for } f_{homeo}(D_{promise}^{exploit}) = 0\\
    a_t^{explor} \sim \pi^{explor}, & \text{for } f_{homeo}(D_{promise}^{exploit}) \not = 0\\
    \end{cases}
%%\]
\end{split}
\end{equation}
%%\end{multline}
where $f_{homeo}(D_{promise})$ with the input of $D_{promise}(t-k,t)^{exploit}$ is Homeo for the decision of exploration mode switching in which its result value are based on Bernoulli distribution and $D_{promise}(t-k,t)^{exploit}$ is the value promise discrepancy of $\pi^{exploit}$. 

%As the pre-training proceeds according to the choice of Homeo,
%\begin{equation}
%\begin{gathered}
%\text{If}\;\;\;s\to\infty,\\\;\;\; 
%\text{then} \lim_{s\to\infty} %J(\theta)_{t}^{exploit} = J(\theta)_{t}^{OPT}     \text{or}    \lim_{s\to\infty} J(\theta)_{t}^{explor} = J(\theta)_{t}^{OPT}    
%\end{gathered}
%\end{equation}
%where  $J(\theta)_{t}^{expoit}$ and $J(\theta)_{t}^{explor}$ are the objective functions of $\pi_{\theta}^{exploit}$ and $\pi_{\theta}^{explor}$ respectively and $J(\theta)_{t}^{OPT}$ is the optimal objective function. 

Then, we expect that the entropy of an Explor agent using $\pi^{explor}$ is greater than that of an Exploit agent using $\pi^{exploit}$ as follows:

\begin{equation}\label{entropy_equation_aps}
%\textit{Uniform\;random} > \textit{PPO} > \textit{TD3}.
\textsc{$\textit{H}\Bigl(\pi^{explor}$}\Bigr) >
\textsc{$\textit{H}\Bigl(\pi^{exploit}$}\Bigr)
\end{equation} where $\textsc{$\textit{H}\Bigl(\cdot$}\Bigr)$ denotes the overall entropy of a policy. %Therefore, we also expect that Explor reach at the best performance faster than Exploit.
Since Exploit using SFs  is used during fine-tuning and Explor takes a supportive role for the exploration of Exploit, $\pi^{explor}$ with $r_{APS}^{exploration}(s, a, s')$ in Eq. \ref{aps_equation} is discarded when a pre-training is finished. Only $\pi^{exploit}$ with $r_{APS}^{exploitation}(s, a, s')$ in Eq. \ref{aps_equation} is used for fine-tuning. 

%$\pi^{Exploit}_{APS}$ denotes an Exploit agent with an intrinsic reward  as $r_{APS}^{exploitation}(s, a, s')$ in Eq. \ref{aps_equation}. $\pi^{Explor}_{APS}$ denotes an Explor agent with an intrinsic reward  as $r_{APS}^{exploration}(s, a, s')$ in Eq. \ref{aps_equation}.} \textcolor{black}{We expect that the entropy of an Explor agent using $\pi^{Explor}_{APS}$ is greater than that of an Exploit agent using $\pi^{Exploit}_{APS}$ as follows:}
%\textcolor{black}{\begin{equation}\label{entropy_equation_aps}
%\textit{Uniform\;random} > \textit{PPO} > \textit{TD3}.
%\textsc{$\textit{H}\Bigl(\pi^{Explor}_{APS}$}\Bigr) >
%\textsc{$\textit{H}\Bigl(\pi^{Exploit}_{APS}$}\Bigr)
%\end{equation}
%where $\textsc{$\textit{H}\Bigl(\cdot$}\Bigr)$ denotes the overall entropy of a policy.}

\subsection{The optimized training method of separate agents} 
In this research, our implementation utilizes an off-policy, DDPG, used as a base algorithm of \cite{84} for all agents. A data from $D_{R}^{\#}$ of either a single agent or both agents can be used to train both agents and determine whether they will synchronize as follows:
%\begin{equation}
%\begin{split}
%Train\_Both\_Agents \bigl( \\    
%&\mkern-18mu\mkern-18mu\mkern-18mu\mkern-18mu\mkern-18mu\mkern-%18mu\mkern-18mu\mkern-18mu\mkern-18mu(..., \textbf{w}, state, %\textbf{c} ...)\sim D_{R}^{exploit}\;'and'\;or\;'or'\; %D_{R}^{explor}\bigr)
%\end{split}
%\end{equation}
%where \textit{Train\_Both\_Agents()} is the pseudocode function of train for both agents and (..., \textbf{w}, state, \textbf{c} ...) is the data loaded from the one of $D_{R}^{x}$ or both of $D_{R}^{x}$ according to the case of 'and' or 'or'.
\begin{equation} \label{train_data}
%\begin{split}
%Train\_Both\_Agents \bigl( \\    
%&\mkern-18mu\mkern-18mu\mkern-18mu\mkern-18mu\mkern-18mu\mkern-%18mu\mkern-18mu\mkern-18mu\mkern-18mu
(..., \textbf{w}, state, ...)\sim D_{R}^{exploit}\;\text{or/and}\; D_{R}^{explor}
%\bigr)
%\end{split}
\end{equation}
where (..., \textbf{w}, state, ...) is the data loaded from  \textcolor{black}{$D_{R}^{\#}$ of both or one of two agents} according to the case of `and' or `or' \textcolor{black}{to be initialized for pre-training and \textbf{w} is a task vector}. In the case of `and' of Eq. (\ref{train_data}), each agent is separately trained with its own $D_{R}^{\#}$. Both agents regarding `or' of Eq. (\ref{train_data}) are trained with one of $D_{R}^{\#}$ in common (see  Table \ref{model_notation} for more details).  

In addition, to maximize the effect of exploration for the model, the  other variance of Explor \textcolor{black}{using} a feature \textcolor{black}{or a skill} can be considered (see  'T. f or s' of  Table \ref{model_notation} for more details). If a feature or skill of Explor is not trained, the feature or skill of Explor with its initial value is used for the agent during pre-training. In the case of feature, the following result of action-value function, $Q$,  can be expected: 
\begin{flalign}
    \begin{aligned}
    %\begin{equation}
    \label{not_trained}
    %\begin{aligned }
    Q^{Explor}_{\phi_{Trained}^{Explor}} \neq Q^{Explor}_{\phi_{NonTrained}^{Explor}}\\
    \end{aligned}
\end{flalign}\textcolor{black}{ where $\phi_{Trained}^{Explor}$ and $\phi_{NonTrained}^{Explor}$ are the trained and the nontrained feature of Explor, respectively. This relies on the fact that Explor will be discarded when the pre-training is finished.}

\begin{algorithm}[!t]
  \caption{The main algorithm of NMPS}
  \label{alg:the_alg}
  \begin{algorithmic}[1]
    %%\Require{\textcolor{black}{$s, g, a_\text{rnvp}$}}
    \Initialize{\strut {Set the value of the duration step of\\\;\;exploration mode}\\
    {Set the value of $\rho$ in (0.1; 0.01; 0.001;\\\;\;0.0001) for Homeo}\\
    \textcolor{black}{
    {Set whether the feature or skill of Explor is\\\;\;or not for \textit{Train\_Both\_Agents}}}
    }
    \Procedure{$\textit{Evaluate}\_\pi^{exploit}$}{$D_{promise}\;\text{of}\;\pi^{exploit}$}
        \State {Compute\;the\;output\;of\;Homeo\;with\;$D_{promise}$\;of\\\;\;\;\;\;\;\;\;$\pi^{exploit}$}
        %%%\State $\;\;with\;D_{promise}\;of\;\pi^{exploit}$
    \EndProcedure
    
    \Procedure{$\textit{Train\_Both\_Agents}$}{$...,\;\textbf{w},\;state, ...$}
        \State {\textcolor{black}{Train\;the\;feature\;of\;$\pi^{exploit}$}}
        \State {\textcolor{black}{Train\;$\pi^{exploit}$\;with\;$r_{APS}^{exploitation}(s, a, s')$}}
        %\State {Train\;$\phi^{explor}(s)$\;according\;to\;the\;condition\;$\textbf{c}$}
        \State {\textcolor{black}{Train\;the\;feature\;or\;skill\;of\;$\pi^{explor}$\;according\;to\\\;\;\;\;\;\;\;\;the\;initialization\;procedure}}
        \State {\textcolor{black}{Train\;$\pi^{explor}$\;with\;$r_{APS}^{exploration}(s, a, s')$}}
    \EndProcedure    
    
	\For {$t=0,\ldots,T-1$}
    	\If{\textit{starting mode}}
                \State {The\;output\;of\;Homeo \;$\gets$ exploration\;mode}
                %%\State {$\;\;\gets exploration\;mode$}
            \Else
                \State {Compute\; $D_{promise}$ \;of\;$\pi^{exploit}$}
                \State {$Evaluate\_\pi^{exploit}(D_{promise}\;\text{of}\;\pi^{exploit})$}
            \EndIf

            \State {$ \textbf{w} \sim$ exploitation\;agent}
            
            %\If{\textit{The output of Homeostasis \\ \qquad\;\; == %exploitation mode} }
            \If{\textit{output of Homeo == exploitation mode} }
                \State {action $\gets \pi^{exploit}$}
            \Else
        	\State {action $\gets \pi^{explor}$}
            \EndIf     
        	\If{\textit{training step of $\pi^{exploit}$ and $\pi^{explor}$}}
                % \State $Train\_Both\_Agents\bigl((..., \textbf{w}, state, \textbf{c} ...)$
                % \State $\;\;\sim D_{R}^{exploit}\;'and' \;or\;'or'\; D_{R}^{explor}\bigr)$
                %
                \State \textcolor{black}{$Train\_Both\_Agents\bigl((..., \textbf{w}, state, ...)$ \\\;\;\;\;\;\;\;\;\;\;\;\;\;$\sim$\; $D_{R}^{exploit}$\;\text{or/and}\; $D_{R}^{explor}\bigr)$}
                %%\State $\;\;\sim D_{R}^{exploit}\;'or'\; or\;'and'\; D_{R}^{explor}\bigr)$                
                %}
            \EndIf
    	\State {Execute action, observe reward (for evaluation) and\\\;\;\;\;\;next\_state}        
	\EndFor        
  %%\algstore{myalg}
  \end{algorithmic}
\end{algorithm}	
% \end{breakablealgorithm}

\subsection{The greater flexibility and generalization of discriminator}
We consider that since Eq. (\ref{eqn:linear_regresssion}) shows that a successor feature and a task vector are a pair to produce a reward, the combination of two factors is crucial to synchronize the two agents. In this research, the decomposition of `task inference' and `exploration' extends to the combination of $\pi^{exploit}$ using SFs and $\pi^{explor}$ of the competence-based algorithms for the purpose of improving exploration performance. The competence-based algorithm maximizes the mutual information $I(\textbf{z};\textbf{w})$ between the encoded observation \textbf{z} and skill \textbf{w} to learn an explicit skill vector as follows:
\begin{equation}
%%\label{eq:valuepromise}
I(\textbf{z};\textbf{w}) = H(\textbf{z}) - H(\textbf{z};\textbf{w}) = H(\textbf{w}) - H(\textbf{w};\textbf{z}).
\end{equation}
\textcolor{black}{Each competence-based algorithm has a different performance in an environment since a competence-based algorithm has a different algorithm based on a different decomposition of the mutual information mentioned above. As a variant of our model, we leverage a competence-based algorithm for exploration. This approach enhances the flexibility and generalization of the Exploit's discriminator, allowing it to overcome the limitations associated with supporting small skill spaces (see $NMPS\_D$ of Table \ref{model_notation}). In the model, the reward of Explor is the reward of DIAYN, which is $NMPS\_D\_sep^{\#\#}$. Action can also be the action of DIAYN instead of the decision of Homeo, which is $NMPS\_D\_sep^{\#\#\_D\_\#\#\#}$.  We also expect the following entropy comparison for each agent of NMPS:}
%\vspace*{-1mm}
\textcolor{black}{\begin{equation}\label{entropy_equation_comp}%\textit{Uniform\;random} > \textit{PPO} > \textit{TD3}.
\textsc{$\textit{H}\Bigl(\pi^{explor}_{competence-based\;algorithm}$}\Bigr) >
\textsc{$\textit{H}\Bigl(\pi^{exploit}$}\Bigr).
\vspace*{-8.5mm}
\end{equation}}\\

\begin{figure*}[!th]
  % \centering
  \begin{subfigure}{.33\textwidth}
    \centering
    \subcaptionbox{Walker}{%
    \includegraphics[scale=0.35]{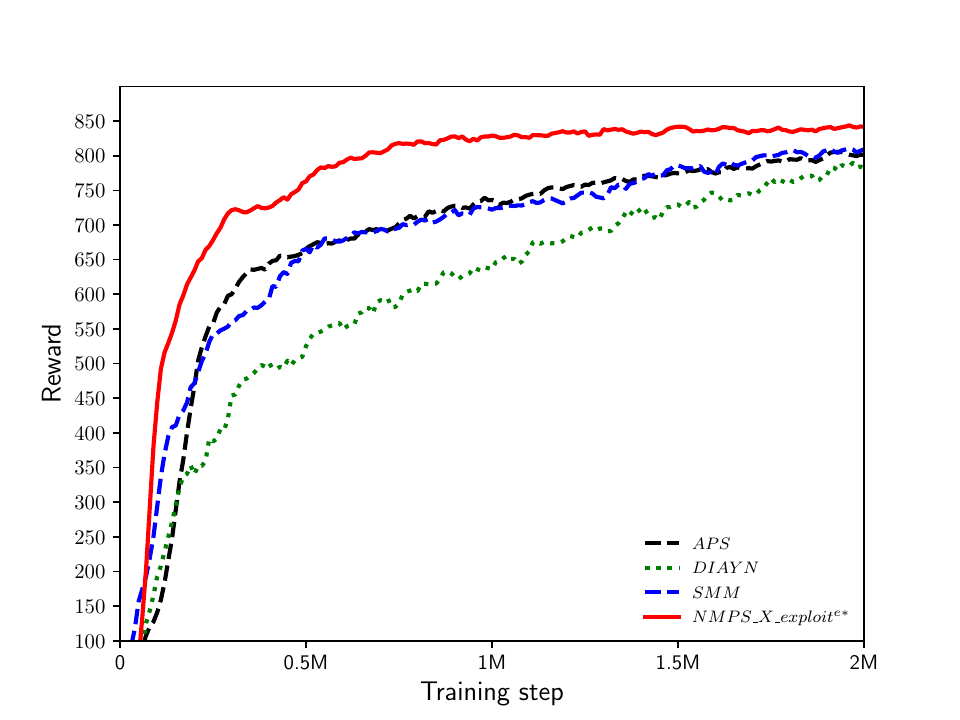}%
    }   
    \label{fig:Walker}
  \end{subfigure}
  \begin{subfigure}{.33\textwidth}
    \centering
    \subcaptionbox{Jaco Arm}{%
    \includegraphics[scale=0.35]{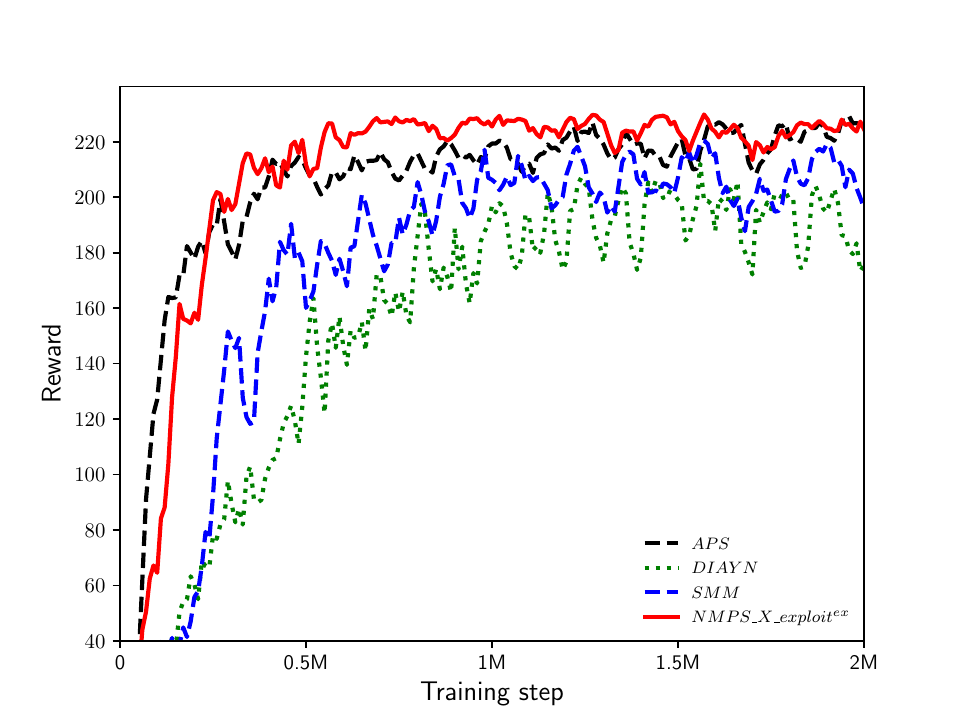}%    
    }       
    \label{fig:Jaco_Arm}
  \end{subfigure}
  \begin{subfigure}{.33\textwidth}   
    \centering
    \subcaptionbox{Quadruped}{%
    \includegraphics[scale=0.35]
    {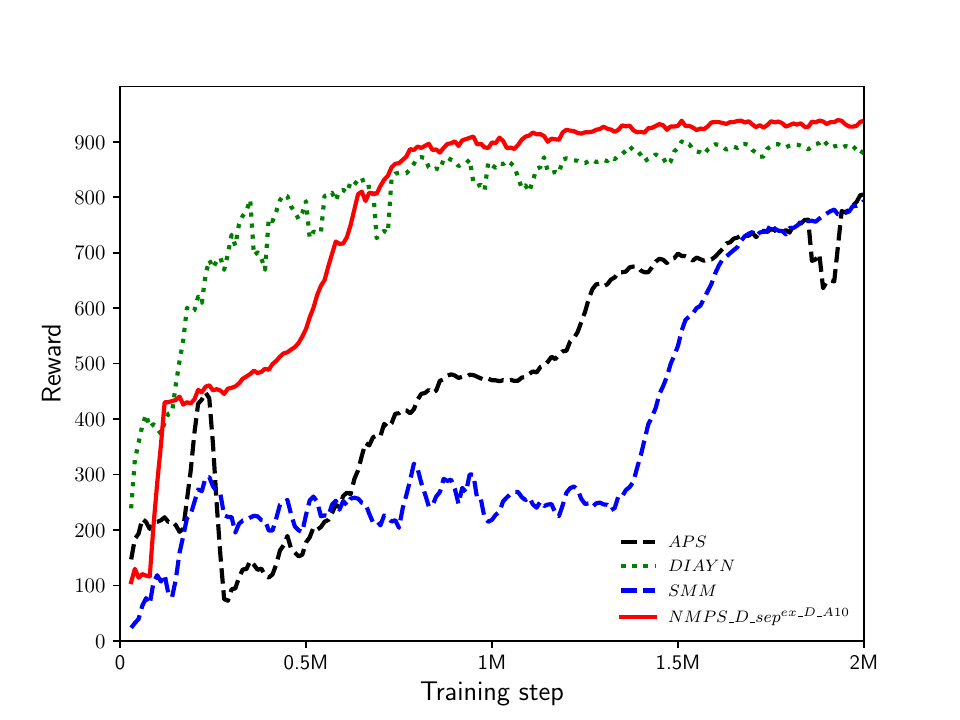}    
    }
    \label{fig:Quadruped2}
  \end{subfigure}
  \caption{The comparison result of fine-tuning of NMPS (the best one), APS, DIAYN and SMM on Walker (a), Jaco Arm (b) and Quadruped (c) by using the intrinsic reward of APS or DIAYN for Explor of NMPS.}
  \label{fig:Result_APS_3_Domains}
%%%\end{figure*}
\end{figure*}

\begin{figure*}[!t]
  \begin{subfigure}{.333\textwidth}
    \centering
    \subcaptionbox{Walker}{%
    \includegraphics[scale=0.35]%{Only_APS_Walker_04132024.pdf}%
    {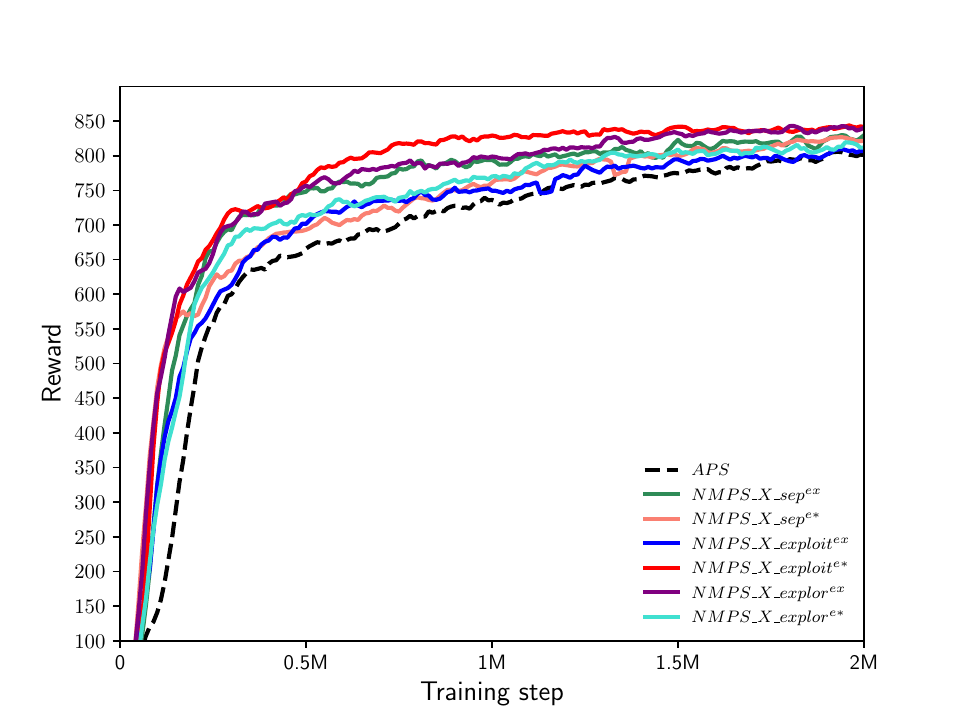}%
    }    
    \label{fig:Ablation_Walker}
  \end{subfigure}
  % \hfill
  \begin{subfigure}{.333\textwidth}
    \centering
    \subcaptionbox{Jaco Arm}{%
    \includegraphics[scale=0.35]%{Only_APS_Jaco_Arm_2403.pdf}%
    {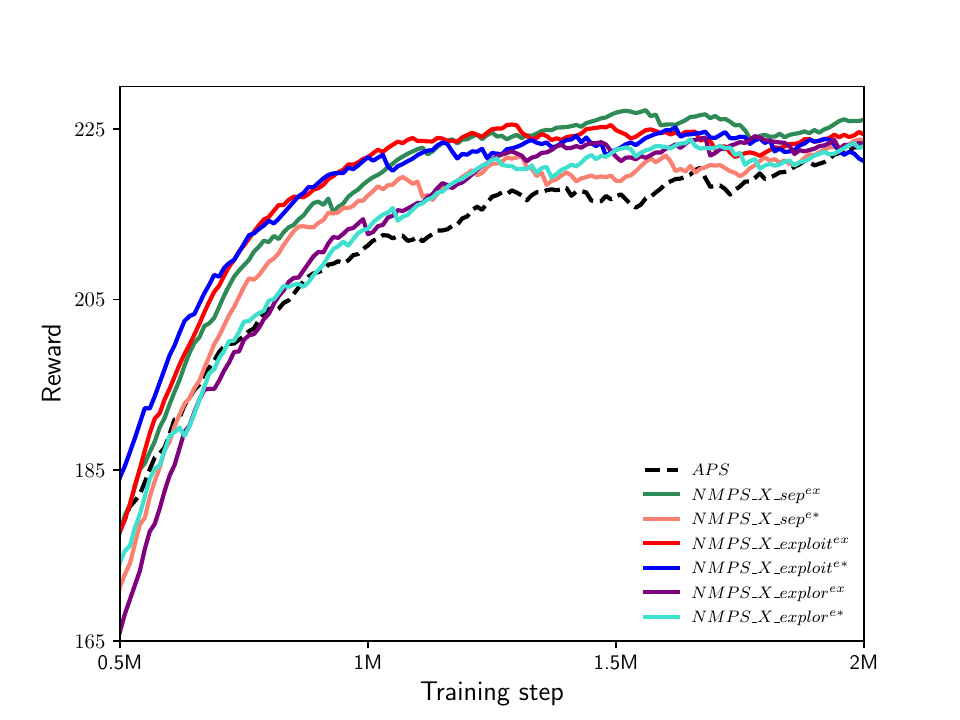}%    
    }       
    \label{fig:Ablation_Jaco_Arm}
  \end{subfigure}
  \begin{subfigure}{.333\textwidth}   
    \centering
   \subcaptionbox{Quadruped}{% 
    \includegraphics[scale=0.35]
    %{quadruped_only_aps_0629_second.pdf}%   
    %{IJCNN_Final_Quadruped_test.pdf}%
    {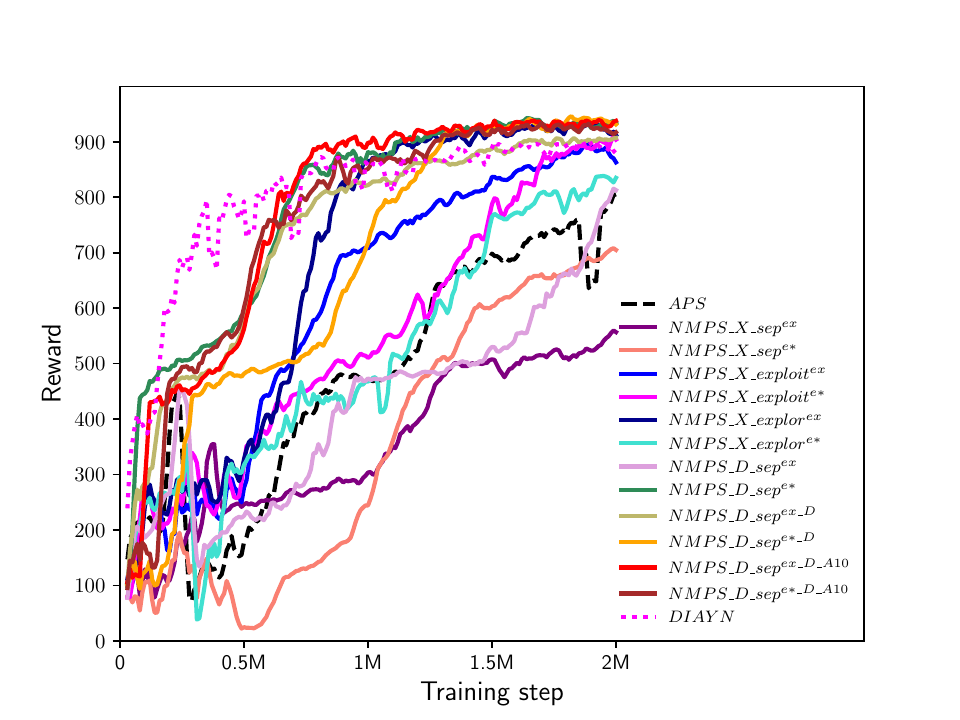}%
    }    
    \label{fig:Ablation_Quadruped}
  \end{subfigure}
  \caption{The comparison result of fine-tuning of all $NMPS\_X$s of NMPS and APS on Walker (a) and Jaco Arm (b, smoothed line) and most NMPS variants and APS on Quadruped (c) by using the intrinsic reward of APS or DIAYN for Explor of NMPS.}
  \label{fig:Ablation_APS_3_Domains}
\end{figure*}

\section{Experiments} \label{Experiments}
The open source  code\footnote{\url{https://github.com/rll-research/url_benchmark}\label{url_benchmark}}  for pre-training and fine-tuning of a competence-based approach evaluated in this research is  used for the comparison, \textcolor{black}{which follows a state-based agent}. Our code is available on the following github\footnote{\url{https://github.com/jangikim2/NMPS}}. The hyper-parameters and Homeo used in this experiment are explained in 'Table I' and 'Algorithm I' of Appendix in 'supplementary document.pdf' on the above github. The evaluation of this research has been conducted in three domains: Walker, Quadruped and Jaco Arm. Three domains are based on the DeepMind Control Suite (DMC), which is a collection of continuous control tasks.

The discrete agents of  $NMPS\_X$ of  our model are based on APS implemented in open source code. Both Exploit and Explor are implemented with $\log q (w | s)$ and $r_{t}^{APT}(s)$ of Eq. \ref{aps_equation}, respectively. That is to say, in $NMPS\_X$, each agent has a separate APS intrinsic reward. Meanwhile, in $NMPS\_D$, DIAYN in competence-based algorithm is used as an Explor instead of Explor of $NMPS\_X$. 

NMPS is implemented on the source code with XI-intra(100, informed, $p^{*}$, X)\footnote{The section 3.1 of \cite{63}} where `XI' means an intrinsic reward, `100' is an explore duration, `informed' is a trigger type, $p^{*}$ is the exploit duration parameterized by a target rate $\rho$ and `X' is the starting mode of explore. Instead of RND used in \cite{63}, this research takes advantage of DDPG since all reference algorithms are based on DDPG in the reference code\textsuperscript{\ref{url_benchmark}}. Therefore, the design of NMPS in this research hinges on off-policy.

Since NMPS is applied to only pre-training, the original code of  fine-tuning in the reference code is utilized for the fine-tuning \textcolor{black}{without a modification}. After fine-tuning, \textcolor{black}{Exploit of} NMPS is compared with APS, DIAYN and SMM \cite{103}. \textcolor{black}{On the other hand, Explor of NMPS is discarded after pre-training.} 

Algorithm \ref{alg:the_alg} shows the explanation of main algorithm of our non-monolithic exploration model during the pre-training phase. During the pre-training phase of $NMPS\_X\_exploit$ or $NMPS\_X\_explor$, the two agents, Exploit and Explor, use a shared replay buffer that is generated by an active agent among the discrete agents. Meanwhile, for pre-training of $NMPS\_D$, separate replay buffers are used for each agent. $NMPS\_D\_sep^{\#\#\_D\_\#\#\#}$ does not need Homeo since only Explor is involved in Action. 

The most important tuned parameter in our source code is the target rate, $\rho$, for Homeo. The best result of each NMPS among 4 values of target rate $\rho$, which are (0.1; 0.01; 0.001; 0.0001), has been selected. The total number of pre-training steps in the reference code is 2M steps. We make a starting mode during 100K steps for exploration by using only the Explor. \textcolor{black}{The snapshot of pre-training made in 1M steps is used for fine-tuning.} A task vector used in both agents is generated by Exploit.

The best results of NMPS and other models (APS, SMM and DIAYN) are compared in Section \ref{best_result}, while all NMPSs and APS are evaluated in Section \ref{Ablation_study}. In Section \ref{four_factors_NMPS}, four factors of NMPS are analysed.

\subsection{Using the intrinsic reward of APS and DIAYN} \label{best_result}
Fig. \ref{fig:Result_APS_3_Domains} presents a comparison of the best performance achieved by NMPS, APS, DIAYN and SMM in all three domains. According to the results, DIAYN and SMM perform worse than or similarly to APS in Walker and Jaco Arm domains. However, in \textcolor{black}{only} Quadruped domain, DIAYN shows remarkable performance in the early stage. Hence, in the experiment on Quadruped domain, NMPS employs DIAYN as Explor \textcolor{black}{for taking advantage of the performance of DIAYN.} Overall, due to the expected effect of aforementioned four factors of NMPS, NMPS shows the best performance among all competence-based algorithms.

% \iffalse
% \subsubsection{Walker : (a) of Figure \ref{fig:Result_APS_3_Domains}}
% $NMPS\_X\_exploit^{e*}$ overwhelms all other performances across all steps. The performance of APS is simialr to that of SMM. DIAYN has the worst performance.

% \subsubsection{Jaco Arm : (b) of Figure \ref{fig:Result_APS_3_Domains}}
% The performance of $NMPS\_X\_exploit^{ex}$ is better than that of APS after around 0.25M steps. SMM and DIAYN (the worst) shows the poor performance over all steps against $NMPS\_X\_exploit^{ex}$ and even APS.

% \subsubsection{Quadruped : (c) of Figure \ref{fig:Result_APS_3_Domains}}
% Although $NMPS\_X\_explor^{ex}$ has a better performance after 1 M steps, DIAYN shows the best performance. The big difference between APS and $NMPS\_X\_explor^{ex}$ is shown after around 0.25M steps. SMM has the worst performance after 0.6M steps.

% $NMPS\_D\_exploit^{e*}$ shows the better performance than $NMPS\_X\_explor^{ex}$ up to around 0.75M. Then, up to around 1.4M, $NMPS\_D\_exploit^{e*}$ is somewhat inferior to $NMPS\_X\_explor^{ex}$. After 1.4M, both models have the same performance.
% \fi
\begin{itemize} 
\item \text{Walker, (a) of Fig. \ref{fig:Result_APS_3_Domains}:} $NMPS\_X\_exploit^{e*}$, which makes the training of feature different from $NMPS\_X\_exploit^{ex}$, outperforms all other models across all steps. The performance of APS is simialr to that of SMM. DIAYN, a skill discovery algorithm, has the worst performance in Walker domain.

\item  \text{Jaco Arm, (b) of Fig. \ref{fig:Result_APS_3_Domains}:} The performance of $NMPS\_X\_exploit^{ex}$ is better than that of APS after around 0.25M steps and is better than those of SMM and DIAYN (the worst) over all steps. SMM and DIAYN exhibit poor performance even when compared to APS. The skill discovery algorithm DIAYN is also not competitive in Jaco Arm.

\item \textcolor{black}{\text{Quadruped, (c) of Fig. \ref{fig:Result_APS_3_Domains}:} Although DIAYN exhibits remarkable performance in the early stage, $NMPS\_D\_sep^{ex\_D\_A10}$ shows the best performance after around 0.68M steps. APS has the third best performance among 4 models. SMM has the worst performance after around 0.6M steps in Quadruped.}
\end{itemize}

In Walker, $NMPS\_X\_exploit^{e*}$ shows overwhelming performance across all steps. It dramatically improves the performance of APS, which is similar to that of DIAYN, since it has three components which are decoupling the combined intrinsic reward of APS, not training feature of Explor and using  $D_{R}^{exploit}$  in common for two agents. The above first two components boosts its exploration by empowering much entropy to Explor. On the other hand, Exploit can hold the privilege of 'task inference' for fine-tuning. Using $D_{R}^{exploit}$  in common can help two agents to synchronize them by using the common data.

In Jaco Arm, the performance of $NMPS\_X\_exploit^{ex}$ is behind in the very early stage. The decoupled agents causes lagging behind APS during the stage. However, after entering the steady state after 0.25M steps, it outperforms APS. Ensuring training feature of Explor makes the stable operation of Explore in the domain.

In Quadruped, $NMPS\_D\_sep^{ex\_D\_A10}$, a variant type, exhibits the best performance in the Quadruped domain, even though $\pi^{explor}_{DIAYN}$ feeds all data into $\pi^{exploit}$ during its pre-training. In the early stage of fine-tuning, its Exploit is behind against DIAYN since it has an exploitation agent, $\pi^{exploit}$, with $r^{exploit}$ during its pre-training. However, during the early stage, the performance of Exploit is better than APS and SMM since the data of $\pi^{explor}_{DIAYN}$ is also in charge of the performance of $\pi^{exploit}$. Although its performance is not enough compared with DIAYN in the early stage, it eventually improves and becomes better than DIAYN, whose performance starts to stabilize after around 0.68M steps. The reason is that SFs of Exploit gives rise to the exact 'task inference' during the steady state of DIAYN.

\subsection{The performance of all NMPSs against that of APS}
\label{Ablation_study}
$NMPS\_X$ in Waker and Jaco Arm are compared with APS. All NMPSs in Quadruped are compared with APS.

\begin{itemize} 
\item \text{Walker, (a) of Fig. \ref{fig:Ablation_APS_3_Domains}:}  $NMPS\_X$ is superior to APS across all steps. The overall performance of each variant type is as follows.\\
\centerline{\hspace{-7mm}\textcolor{black}{$NMPS\_X\_explor > NMPS\_X\_exploit >$}} \\ \centerline{\hspace{-7mm}\textcolor{black}{$NMPS\_X\_sep > APS$}}
\item  \text{Jaco Arm, (b) of Fig. \ref{fig:Ablation_APS_3_Domains}:} Since the performance difference between $NMPS\_X$ and APS is small, the graph shows a smoothed line. However, all variants of $NMPS\_X$ exhibit better performance than APS after at least 0.7M steps. The overall performance of each variant type is as follows.\\
\centerline{\hspace{-7mm}\textcolor{black}{$NMPS\_X\_exploit > NMPS\_X\_sep >$}} \\ \centerline{\hspace{-7mm} \textcolor{black}{$NMPS\_X\_explor > APS$}}
\item \text{Quadruped, (c) of Fig. \ref{fig:Ablation_APS_3_Domains}:} Most NMPS variants are superior to APS. $NMPS\_D\_sep^{\#\#\_D\_A10}$ shows the best performance. The overall performance order of each variant is as follows.\\
\centerline{\hspace{-7mm}\textcolor{black}{$NMPS\_D\_sep^{\#\#\_D\_A10} > NMPS\_D\_sep^{\#\#\_D} >$}} \\
\centerline{\hspace{-7mm}\textcolor{black}{$NMPS\_X\_explor > NMPS\_D\_sep  >$}} \\ \centerline{\hspace{-7mm}\textcolor{black}{$  NMPS\_X\_exploit > APS > NMPS\_X\_sep$}}
\end{itemize}

In Walker, $NMPS\_X$ has the strength against APS. Especially, in this domain, the performance of $NMPS\_X\_sep$ is behind against that of other types, $NMPS\_X\_explor$ and $NMPS\_X\_exploit$.

In Jaco Arm, even though the difference of performance between $NMPS\_X$ and APS is small, all $NMPS\_X$s are superior to APS. The performance of $NMPS\_X\_explor$ is behind that of other two types, $NMPS\_X\_exploit$ and $NMPS\_X\_sep$.

In Quadruped, although most $NMPS\_X$ variants are superior to APS, $NMPS\_D$ improves the performance of $NMPS\_X$ further. The difference between the performance of $NMPS\_D$ and that of $NMPS\_X$ is big. In the performance difference, the potential of $NMPS\_D$ is remarkable for how to use an agent with SFs. $NMPS\_D\_sep^{\#\#\_D\_\#\#\#}$ shows high performance among all NMPSs. Among them, the performance of $NMPS\_D\_sep^{\#\#\_D\_A10}$ is highest. We pay attention to the dimension of feature and skill of $NMPS\_D\_sep^{\#\#\_D\_\#\#\#}$. In addition, it is noticeable that the performance of $NMPS\_D\_sep^{e*}$ is better than that of $NMPS\_D\_sep^{\#\#\_D}$. If the dimension of feature and skill of NMPS is not the same as that of practically optimized feature and skill such as $NMPS\_D\_sep^{\#\#\_D\_A10}$, the performance of $NMPS\_D\_sep^{\#\#\_D}$ is not enough to overcome even that of $NMPS\_D\_sep^{e*}$.

\subsection{The analysis of four factors of NMPS} \label{four_factors_NMPS}

Although NMPS using decoupled intrinsic rewards gives rise to the performance improvement of APS, $NMPS\_X\_sep$ is not enough to overcome APS as shown in Quadruped of Fig. \ref{fig:Ablation_APS_3_Domains}.

In Walker and Jaco Arm, $NMPS\_X\_exploit$ verifies a potential competitiveness  across two domains. Furthermore, it is noteworthy that synchronizing both Exploit and Explor through  $NMPS\_X\_exploit$ and $NMPS\_X\_explor$ is significant to boost the performance of normal type, $NMPS\_X\_sep$.

The effect of the factor regarding "The train of feature or skill" depends on the domain, as the purpose of this factor introduces a significant variance into NMPS.

Based on the performance results in Quadruped of Fig. \ref{fig:Ablation_APS_3_Domains}, an agent with SFs can overcome the limitation of supporting only small skill spaces of a discriminator by leveraging the superior performance of another competence-based algorithm in the environment. 

\section{Discussion} \label{Discussion}
\subsection{Decoupling a monolithic exploration agent with SFs}
The variants of NMPS demonstrate the contribution of NMPS. Most NMPS variants outperform APS in all environments. Decoupling exploitation and exploration in the original monolithic algorithm with SFs improves its performance. The mode-switching controller (Homeo in this research) for both agents impacts the rate of communication with the environment. As a result, agents with SFs based on NMPS show better performance than APS during fine-tuning.

\subsection{The factors to maximize the performance of NMPS}
In this research, several factors are found to influence the performance of decoupled unsupervised pre-training. These factors include \textcolor{black}{the factor of mode switching controller, i.e. the target rate $\rho$,} and four factors related to the variants of NMPS. The Homeo controller's target rate, $\rho$, is found to be a crucial factor. Furthermore, \ref{Ablation_study} shows that the performance difference among the variants of $NMPS\_X$ and $NMPS\_D$ are due to the four factors related to their design. These factors result in performance differences among the four groups of NMPS across different environments. Notably, even though there are the agents of same type, taking advantage of {$D_{R}^{\#}$} and $'T.\;f\;or\;s'$ makes a different performance. They are other breakthroughs for $NMPS\_X$ and $NMPS\_D$ to overcome their own limitation.

\subsection{The vulnerable point of competence-based approach}
We believe that the performance of DIAYN in Quadruped domain is better than that of APS due to the feature dimension of APS to some extent since the feature dimension of APS, 10, is smaller than the skill dimension of DIAYN which is 16.

Interestingly, the performance of $NMPS\_D\_sep^{\#\#\_D\_A10}$ based on the feature dimension of APS is better than that of $NMPS\_D\_sep^{\#\#\_D}$ and DIAYN. The extreme non-monolithic operation of Explor of $NMPS\_D\_sep^{\#\#\_D\_\#\#\#}$ has the full charge of exploration as well as exploitation. Finally, $NMPS\_D\_sep^{\#\#\_D\_A10}$ suggests a new methodology in order to overcome the linear reward-regression issue of agent using SFs.

\section{Conclusion} \label{Conclusion}
In summary, this research proposes a novel unsupervised pre-training method with SFs, NMPS. To improve performance, it incorporates a non-monolithic exploration methodology with greater flexibility, generalization of the discriminator, and optimized training methods for each agent's feature or skill.

Further research is needed to enhance our model for an agent which is not based on features or skills for Explor.

\bibliographystyle{IEEEtran} 
\bibliography{IJCNN_2024_Final}

\end{document}